%% file: main.tex
\pdfoutput=1
\documentclass[sigconf, review=false]{acmart}

\usepackage{booktabs} 
\usepackage{subcaption}
\usepackage{graphicx} 
\usepackage{tabularx} 
\usepackage[hide]{notes-alt}

\usepackage{amsmath}
\usepackage{mathtools}
\usepackage{amsfonts}
\usepackage{algorithm}
\usepackage[noend]{algpseudocode}
\usepackage{xspace}
\usepackage{enumitem}

\newtheorem{theorem}{Theorem}
\newtheorem{hypothesis}{Hypothesis}
\newcommand{\problem}{\textsf{ALiR}\xspace}

\newcommand{\baseline}{\texttt{Hogwild}\xspace}
\newcommand{\mllib}{\texttt{MLLib}\xspace}

\newcommand{\mr}{\textsc{MapReduce}\xspace}

\newcommand{\men}{\textsf{MEN}\xspace}
\newcommand{\rg}{\textsf{RG65}\xspace}
\newcommand{\google}{\textsf{Google}\xspace}
\newcommand{\ap}{\textsf{AP}\xspace}
\newcommand{\ws}{\textsf{WS353}\xspace}
\newcommand{\rw}{\textsf{RareWords}\xspace}
\newcommand{\battig}{\textsf{Battig}\xspace}
\newcommand{\semeval}{\textsf{SemEval}\xspace}

\newcommand{\concat}{\textsc{Concat}\xspace}
\newcommand{\pca}{\textsc{Pca}\xspace}

\newcommand{\E}{\mathbb{E}}
\newcommand{\w}{\mathbf{w}}

\newcommand{\mpara}[1]{\medskip\noindent{\bf #1}}

\newcommand{\wordtovec}{\textsc{Word2Vec}\xspace}
\newcommand{\partition}{\textsc{equal partitioning}\xspace}
\newcommand{\shuffle}{\textsc{Shuffle}\xspace}
\newcommand{\random}{\textsc{random sampling}\xspace}
\providecommand{\norm}[1]{\lVert#1\rVert}

\definecolor{linkcolor}{rgb}{0.7,0,0}
\definecolor{citecolor}{rgb}{0,0.5,0}
\definecolor{urlcolor}{rgb}{0,0,0.7}

\begin{document}

\fancyhead{}

\copyrightyear{2019}
\acmYear{2019}
\setcopyright{acmcopyright}
\acmConference[WSDM '19]{The Twelfth ACM International Conference on Web
Search and Data Mining}{February 11--15, 2019}{Melbourne, VIC,
Australia}
\acmBooktitle{The Twelfth ACM International Conference on Web Search and
Data Mining (WSDM '19), February 11--15, 2019, Melbourne, VIC,
Australia}
\acmPrice{15.00}
\acmDOI{10.1145/3289600.3291011}
\acmISBN{978-1-4503-5940-5/19/02}
\title{Asynchronous Training of Word Embeddings for Large Text Corpora}

\author{ Avishek Anand }
\affiliation{%
  \institution{L3S Research Center}
  \streetaddress{Appelstra{\ss}e 4}
  \city{Hannover}
  \state{Lower Saxony}
  \postcode{30167}
  \country{Germany}
}
\email{anand@l3s.de}
\author{Megha Khosla}
\affiliation{%
  \institution{L3S Research Center}
  \streetaddress{Appelstra{\ss}e 9a}
  \city{Hannover}
  \state{Lower Saxony}
  \postcode{30167}
  \country{Germany}
}
\email{khosla@l3s.de}
\author{Jaspreet Singh}
\affiliation{%
  \institution{L3S Research Center}
  \streetaddress{Appelstra{\ss}e 4}
  \city{Hannover}
  \state{Lower Saxony}
  \postcode{30167}
  \country{Germany}
}
\email{singh@l3s.de}
\author{Jan-Hendrik Zab}
\affiliation{%
  \institution{L3S Research Center}
  \streetaddress{Appelstra{\ss}e 4}
  \city{Hannover}
  \state{Lower Saxony}
  \postcode{30167}
  \country{Germany}
}
\email{zab@l3s.de}
\author{Zijian Zhang}
\affiliation{%
  \institution{L3S Research Center}
  \streetaddress{Appelstra{\ss}e 9a}
  \city{Hannover}
  \state{Lower Saxony}
  \postcode{30167}
  \country{Germany}
}
\email{zzhang@l3s.de}


\renewcommand{\shortauthors}{A. Anand et al.}

\begin{abstract}
Word embeddings are a powerful approach for analyzing language and have been widely popular in numerous tasks in information retrieval and text mining. 
Training embeddings over huge corpora is computationally expensive because the input is typically sequentially processed and parameters are synchronously updated. Distributed architectures for asynchronous training that have been proposed either focus on scaling vocabulary sizes and dimensionality or suffer from expensive synchronization latencies.

In this paper, we propose a scalable approach to train word embeddings by  partitioning the \emph{input space} instead in order to scale to massive text corpora while not sacrificing the performance of the embeddings. Our training procedure does not involve any parameter synchronization except a final sub-model merge phase that typically executes in a few minutes. 
Our distributed training scales seamlessly to large corpus sizes and we get comparable and sometimes even up to 45\% performance improvement in a variety of NLP benchmarks using models trained by our distributed procedure which requires $1/10$ of the time taken by the baseline approach. Finally we also show that we are robust to missing words in sub-models and are able to effectively reconstruct word representations.



\end{abstract}




\maketitle 

\input{intro-new}
\input{related-work}

\input{approach}
\input{evaluation}

\input{conclusions}

\begin{acks}
This work is partially funded by ALEXANDRIA (ERC 339233) and SoBigData (Grant agreement No. 654024).
\end{acks}


\bibliographystyle{abbrv}
\bibliography{references} 
\end{document}

%% file: intro-new.tex
\section{Introduction}
\label{sec:intro}
Word representations or word embeddings are low dimensional, continuous and dense representations of words in a semantic vector space. Such embeddings are routinely used as input representations in neural network architectures tasks such as syntactic parsing \cite{socher2013parsing}, sentiment analysis \cite{socher2013recursive}, machine translation \cite{cho2014learning}, image annotation \cite{weston2011wsabie}, query modeling, document representations~\cite{zamani2016estimating} and in the construction of neural models for ranking~\cite{mitra2017neural,desm16}.


Word representations are typically learned in an unsupervised manner from large text corpora -- traditionally by matrix factorization approaches~\cite{li2015word} and more recently by employing neural networks~\cite{mikolov2013distributed,mikolov2013linguistic,bojanowski2016enriching}. One particularly popular implementation is \emph{skip-gram with negative sampling} (SGNS)~\cite{mikolov2013linguistic} also known as \textsc{Word2Vec}. The popularity of the SGNS approach is due to faster training times based on the advancements in asynchronous gradient descent through lock-free updates~\cite{recht2011hogwild} and careful model updates using negative sampling~\cite{ji2016parallelizing}, sub-sampling and vocabulary pruning~\cite{mikolov2013distributed}. Although these improvements make instance-level training faster, the input inherently has to be processed sequentially making training massive datasets slower. As an example \emph{training embeddings over the entire English Wikipedia ($\approx 14$ GB) takes around 17.8 hours with all the optimizations described}.

Distributed architectures for SGNS have also been proposed to this extent but with different objectives. Ordentlich et al.~\cite{ordentlich2016network} try scaling SGNS training by partitioning the embedding dimensions, while~\cite{stergiou2017distributed} partition the vocabulary space to scale to large vocabulary sizes. We propose approaches that instead partition the input space. Other distributed approaches for training ML models~\cite{ghoting2011systemml,zaharia2010spark,boehm2014hybrid,xing2015petuum} that do partition the input space rely on expensive parameter synchronization procedures that typically incur network latencies. 

In this paper, we propose a scalable approach to train word embeddings by  partitioning the \emph{input space} in order to scale to massive text corpora while not sacrificing the performance of the embeddings. Our approach is simple, easy to implement and effective when compared against both centralized and distributed baseline approaches in a host of word similarity, analogy and categorization benchmarks. Our training procedure does not involve any parameter synchronization except a final sub-model merge phase that typically executes in a few minutes.

In detail, we propose a simple data-division strategy by creating representative random samples from the input data. We support our data-division strategy using the result from~\cite{levy2014neural} which shows that for a \wordtovec model trained using negative sampling, the representational capacity of trained embeddings depends on the word and word-context distributions. We show that our sampling techniques in fact preserve these distributions, which allows us to build sub-models without deviating much from the original performance of the word representations.
More importantly, this enables our models to be trained in parallel and asynchronously over each of these samples without the need of any parameter synchronization. 
 Our sampling strategy therefore offers us the following two main advantages over previous work in this direction.
Firstly, it allows our approach to be truly parallel depending on the number of existing workers and is independent of the skew and embeddings as in~\cite{ordentlich2016network} and \cite{stergiou2017distributed} respectively. Training can therefore be easily leveraged by highly scalable parallel data processing platforms. Secondly, and more importantly there are no intermediate synchronization phases eliminating further network latencies.

In the end a combined model is generated by merging all asynchronously trained representations into a single consistent representative embedding that actively takes sparsity and missing vocabulary into account. We propose a generalization of Generalized Procrustes Analysis to merge sub-models with missing vocabulary.

Finally, we perform extensive evaluation on two large text datasets Wikipedia (14 GB) and Web (268 GB) to demonstrate the effectiveness and efficiency of our approaches. Our distributed training scales seamlessly to the corpus sizes and we show that \emph{we can train embeddings on the Wikipedia dataset in 2 hours as opposed to 17.8 hours} (with the standard implementation inclusive of asynchronous SGD and negative sampling). Moreover, we show that we get comparable and sometimes even up to 45\% performance improvement in a variety of NLP benchmarks using models trained by our distributed procedure which requires $1/10$ of the time taken by the baseline approach. Finally we show that we are robust to missing words in sub-models and are able to effectively reconstruct word representations. We release the code for our implementation in \texttt{\url{https://github.com/jhzab/dist_w2v}}.

In summary, we make the following contributions in this paper:

\begin{itemize}
\item We propose asynchronous methods for training word embedding models on large input text corpora based on simple, easy to implement sampling approach supported by empirical and theoretical justifications.


\item We propose effective sub-model merging approach \problem that is also robust to out-of-vocabulary terms.


\item We perform extensive experimental evaluation on 14 GB of a Wikipedia and 268 GB of Web to showcase both the scalability and effectiveness of our approaches.
\end{itemize}


%% file: related-work.tex
\section{Related Work}

We classify the previous works into the following categories: (1) parallelizing SGD given large input corpora, (2) learning reliable embeddings by combining word representations.



\textbf{Efficient and Scalable SGNS.} 
The original implementation of \wordtovec by  Mikolov et al.~\cite{mikolov2013distributed}
uses Hogwild~\cite{recht2011hogwild} to parallelize stochastic gradient descent (SGD). Hogwild is a parallel
SGD  algorithm  that  seeks  to  ignore  conflicts  between  model
updates on  different  threads  and  allows  updates  to  proceed
even in the presence of conflicts. Indeed for large vocabulary sizes updates across threads are unlikely to be of the same input word which explains the rarity of conflicts which could be ignored without affecting the convergence. 
Popular implementation of \wordtovec, later implemented into software packages like Gensim
~\footnote{http://rare-technologies.com/word2vec-in-python-part-two-optimizing/}, 
TensorFlow
~\footnote{https://www.tensorflow.org} 
uses multithreading to increase training speed. 


The approach by Ji et al. \cite{ji2016parallelizing}, implemented similarly in MLlib~\footnote{https://spark.apache.org/docs/latest/mllib-feature-extraction.html}, further exploits the locality in model updates by combining the lock-free scheme of Hogwild with mini-batching of model updates involving the same target word and shared negative samples. Their shared memory multi-core solutions are thereby able to exploit level-3 BLAS operations. However, their distributed implementation is still not perfectly asynchronous and needs global parameter synchronizations.

Vuurens et.al.~\cite{eickhoff2016efficient} on the other hand proposes an efficient caching strategy that provides a comparable efficiency gain over the hierarchical softmax variant of the \wordtovec. Ordentlich et. al \cite{ordentlich2016network} proposes a distributed  \wordtovec training procedure that distributes the word vectors by partitioning the embedding dimensions across workers and parallelizes vector training to reduce training latency. Unlike, partitioning the embedded dimensions~\cite{ordentlich2016network} we draw multiple samplefrom the training set. Our approach is naturally scalable to increasing training data sizes unlike embedding partitioning where the scalability is limited to the embeddings dimensionality.



Recently, Stergiou et al.~\cite{stergiou2017distributed} also propose a partitioning approach where the objective is to scale to large vocabularies. Specifically they develop a distributed algorithm for sampling from a discrete distribution and use it to optimize Negative Sampling for SGNS \wordtovec which allows scaling to large vocabularies. We note that our focus is different from this work as we aim to scale for large training sets instead of large vocabularies.




\textbf{Combining Word Representations.} Now we review the approaches that have been employed to merge multiple trained models. Garten et al.~\cite{garten2015combining} put forward an approach, where a model from \wordtovec and a model from DVRS~\cite{ustun2014distributed} are combined. As a combination strategy, they employ vector concatenation and linear addition between vectors corresponding to the same word from different models and demonstrate  that the combined model performs even better than the best setting of individual ones, especially when the training data is limited.
For learning reliable embeddings from a limited training data Avo Murom{\"a}gi et al. \cite{muromagi2017linear} propose a different strategy such that they combine the models trained with the same system and on the same dataset, albeit using different random initialization.  Goikoetxea et al. \cite{goikoetxea2016single} show that a simple concatenation of independently learned embeddings from different sources like text corpora or WordNet outperforms more complex combination techniques in word similarity and relatedness datasets. We use this as our baseline as well referred to as \concat. Ghannay et al.~\cite{ghannay2016word}  evaluates different approaches to combine word embeddings to identify effective word embeddings that can achieve good performances on all tasks. None of these approaches take into account missing word information, or out-of-vocabulary (OOV) terms, and have a strong assumption that all the input embeddings should have the same vocabulary. 
Speer and Chin \cite{DBLP:journals/corr/SpeerC16} present an ensemble  method  that  combines  embeddings  produced  by  GloVe \cite{Pennington2014glove} and \wordtovec with structured knowledge sources,  merging  their  information  into  a
common  representation  with  a  large,  multilingual  vocabulary. They use a locally linear alignment procedure \cite{inproc:Zhao2015} to align overlapping words in GloVe and \wordtovec embeddings. The embedding corresponding to a non overlapping word is then computed  as the average of the
embeddings of the nearest overlapping terms,  weighted by their cosine similarity.
The other work that attempts to reconcile OOV information is~\cite{yin2015learning} but our merging approach \problem can be seen as a generalization of their approach where their result is the output after one round of \problem.

%% file: approach.tex
\section{Our Approach}

\label{sec:appr}
\textbf{Preliminaries.} In this work, we focus on the SGNS implementation of \wordtovec. SGNS assumes a collection of words $\{w_i\} \in V_W$ and their contexts $c\in V_C$, where $V_W$ and $V_C$ are the word and context vocabularies. A \emph{context} $c\in V_C$ of the word $w_i$ is a word from the sequence of words $w_{i-win},..., w_{i-1},w_{i+1},...,w_{i+win}$ for some fixed window size $win$. Let $D$ be a multi-set of all word-context pairs observed in the corpus.
Let $\vec{w}, \vec{c} \in \mathbb{R}^d$ be the $d$-dimensional \textit{word embeddings} of word $w$ and context $c$, specified by the mappings: $\mathcal{W}: V_W \to \mathbb{R}^d,\quad \mathcal{C}: V_C \to \mathbb{R}^d.$
SGNS aims to find mappings $\mathcal{W}$ and $\mathcal{C}$ such that the following objective specified for each pair $(w,c) \in D$ is maximized.
 \begin{equation}
\label{eq:objective_expec_full}
\log\sigma(\vec{w}. \vec{c}) + k\cdot\mathbb{E}_{c' \sim P_D}(\log\sigma(- \vec{w}. \vec{c'})) ,
\end{equation}
where $\sigma(x) = {1\over 1-exp(-x)}$ is the sigmoid function. For each positive example $(w,c)$, $k$ negative samples $(w,c')$ are drawn from a noise distribution. Here we use the unigram distribution raised to power of $3/4$ as the noise distribution over contexts similar to the original paper \cite{mikolov2013distributed}. Usually Equation \eqref{eq:objective_expec_full} is optimized via the stochastic gradient descent procedure that is performed during passing through the corpus \cite{mikolov2013distributed}.

Previous work has also focused on the theoretical analysis of SGNS, for instance, Levy and Goldberg~\cite{levy2014neural} showed that for sufficiently large $d$ and generating negative samples via uniform distribution over unigrams, SGNS is an implicit matrix factorization of shifted PMI matrix. In particular, they showed that SGNS's objective is optimized by setting $\vec{w}\cdot\vec{c} = PMI(w,c)-\log k$ for every $(w, c)$ pair, i.e,
\begin{equation}
\vec{w}\cdot\vec{c}= \log {P(w,c)\over P(w)P(c)} -\log k,
\label{eq:gl}
\end{equation}
where $P(w,c)$ is the joint probability distribution of word context pairs, $P(w)$ and $P(c)$ are the probability distributions for word and context respectively in the given text corpora. 

The above results suggests that if we train SGNS models on two corpora with the same word (unigram) and word-context (bigram) distributions we would expect similar embeddings which also motivates our approach as outlined below.


\label{sec:approach}

\begin{figure}[t!]
\centering
     \begin{subfigure}{.4\linewidth}
     	\includegraphics[width=\linewidth]{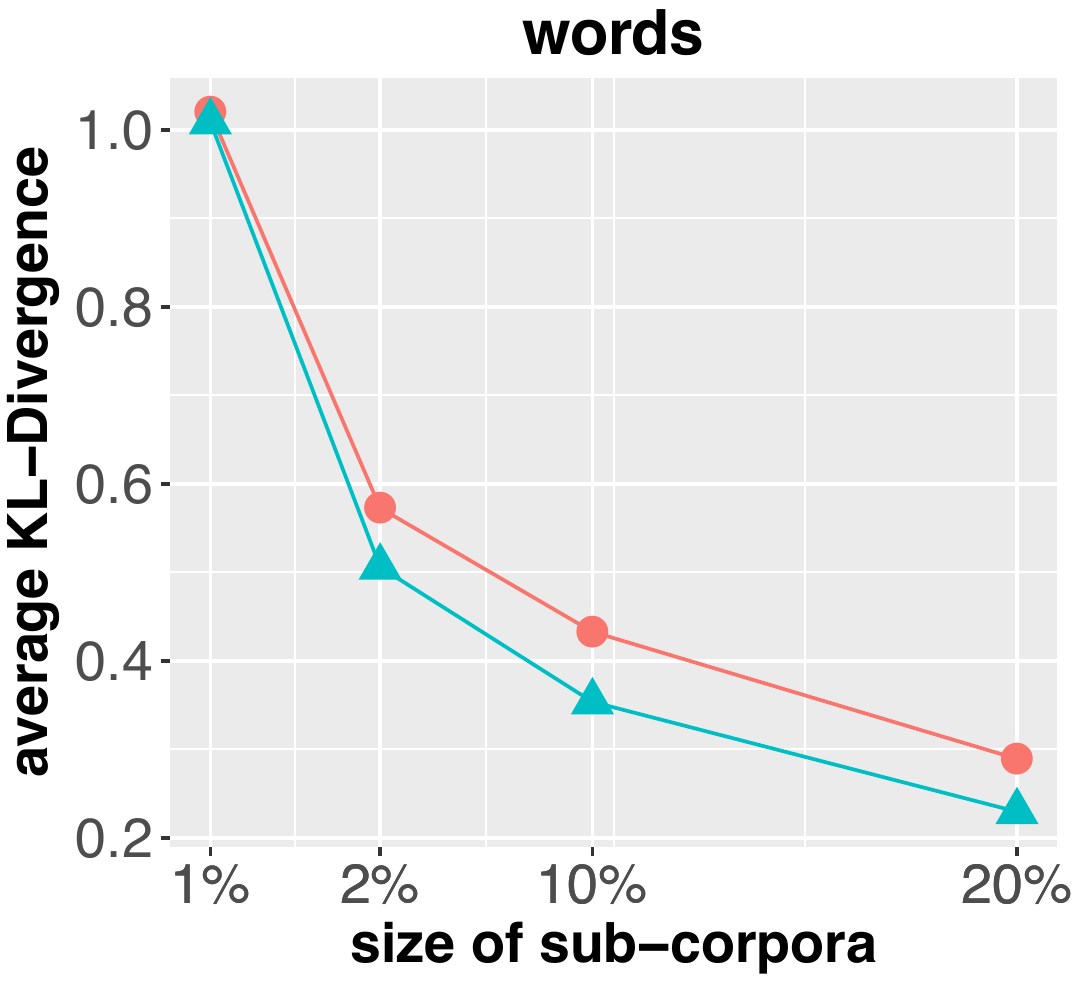}
    \end{subfigure}
    \begin{subfigure}{.4\linewidth}
		\includegraphics[width=\linewidth]{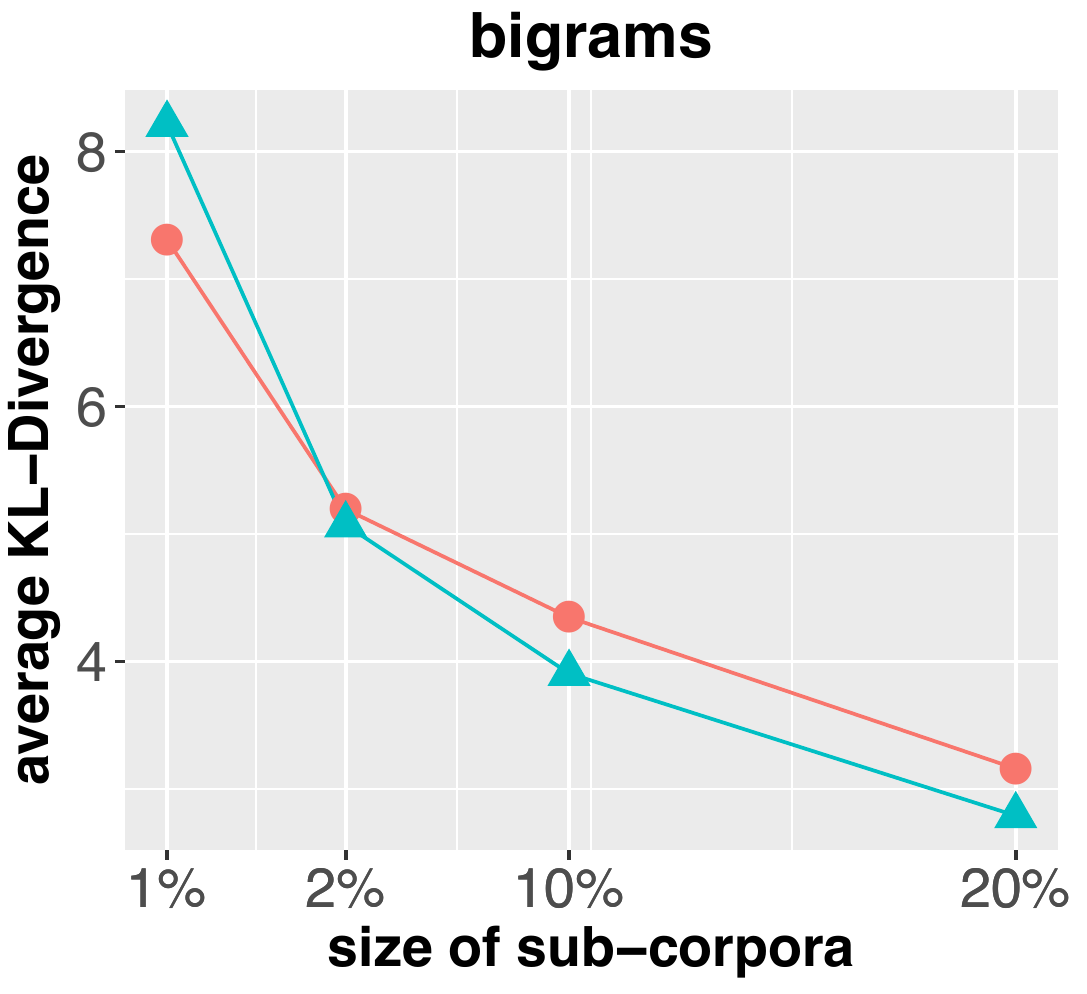}
	\end{subfigure} 
    	\caption{Average KL-Divergence of words/bigrams distribution from sub-corpus to original corpus, averaged over 10 sub-corpora. (partitioning in Red, and Random Sampling in Blue)}
        \label{fig:freq}
\end{figure}
\mpara{Objective and Approach Outline.} 
Our approach is built on the following hypothesis which we elaborate and validate in the rest of the paper. 

\begin{hypothesis} \label{hyp:1}
Let the original corpus $~C$ is divided into sub-corpora $~C_1, C_2, \cdots , C_k$ such that the unigram and bigram distributions are preserved (on average) with respect to their distributions in the original corpus. The final word representations for $C$ (with comparable quality) can then be obtained by finding the combined representations of the asynchronously trained embeddings for the sub-corpora.
\end{hypothesis}

In order to validate our hypothesis we begin by proposing a easy to implement random sampling scheme (supported by theoretical and empirical justifications) to \emph{divide} the original input corpus, in the \textbf{divide phase} (cf Section~\ref{sec:divide}), into multiple sub-corpora such that each sub-corpus is an independent representative sample of the original input. The independence criteria allows us to train sub-corpora asynchronously in parallel, in the \textbf{train phase} (cf Section~\ref{sec:map-reduce}), which leads to the speedups in training time. The second part of our hypothesis deals with finding a \emph{combined representation} from the obtained representations of the sub-models in the \textbf{merge phase} (cf Section~\ref{sec:merge-phase}). In addition to testing some simple schemes of finding the combined representation we develop a variant of Generalized Procrustes Analysis\cite{Gower1975} which helps to find effective combined representation over the union of vocabulary of all sub-models (note that some words might be missing in some sub-models).

\subsection{The Divide Phase}
\label{sec:divide}
The first part of our hypothesis deals with the divide phase where our proposed strategy should divide the data into a number of sub-corpora such that the unigram and bigram distributions are preserved. 
From the distributional hypothesis we know that the quality of a word embedding suffers if a large fraction of its context/co-occuring words are missing from the sub-corpus $C'$. In other words it is desirable to ensure that we do not miss words and word contexts in a sub-corpora.


Based on this intuition, we propose a simple and effective random sampling approach to divide the input data (a set of sentences) into multiple smaller sub-corpora. In particular we propose \random using a sampling rate of $r(\text{in} \%)$, in which we choose $100/r$ samples of $rN/100$ sentences each by choosing sentences independently and uniformly at random \emph{with replacement}. We support our proposed technique by showing theoretically and empirically that the unigram and bigram distributions are preserved, on average, in the sub-corpora. In particular we prove the following theorem in case of unigram distribution. 

\begin{theorem}
Let $|\w|_C$ denote the frequency of word $\w$ in a given corpus $C$ and ${T}_{C}$ denote the total number of words in $C$ . Let $P_C(\w)$ denotes the probability that a randomly drawn out word from any sentence in $C$ is the word $\w$. For any sub-corpus $C'$ drawn from the original corpus $C$ using \random and any word , $\w$ we have 
$\E\left( |\w|_{C'} \over \mathcal{T}_{C'} \right) = P_C(w)$.
\end{theorem} 

We also analytically compute a threshold such that if the probability of occurrence of a word $\w$ in $C$ is above this threshold, then the expected number of sub-corpora not containing $\w$ is exponentially small in $N$.
\begin{theorem}
Let $u=r/100$ where $r$ is the sampling rate. Let $\ell$ be the sentence length. If $P_C(w)> 1-(1-u)^{1-u \over \ell u}, $
then the expected number of randomly sampled sub-corpora which misses a word $\w$ is $\exp(-O(N))$.
\end{theorem}
Plugging in for example $u=0.1$ and $\ell=100$, we infer from the above result that it is highly unlikely that a word with an occurrence probability greater than $0.0095$ is missed. We also show empirically that our sampling strategy allows us to cover a large percentage of the vocabulary.
Because of space limitations, the proofs are provided in the supplementary material.





\mpara{Empirical Evidence.} In Figure~\ref{fig:freq} we plot the KL divergence from the empirical unigram and bigram distributions of a sample (averaged by randomly picking $10$ samples from all samples or partitions) to distributions of the complete training data. In this experiment we compare \random with another simple data division approach, called \partition, in which we sequentially divide the whole corpus containing $N$ sentences into $100/r$ partitions containing equal number of $rN/100$ sentences.
The lower KL divergence in the unigram and the bigram distribution using \random strategy indicates that the randomly created samples are better representatives of the complete training data as compared to those created by the \partition strategy. In addition we also show that each created sub-model corresponding to a small sampling rate (see Section ~\ref{sec:sampling}) of input data performs comparable to the model built on the complete corpus.
We also present the statistics of vocabulary sizes of the original corpus and that covered by the sampled sub-corpora in the supplementary material. In particular, \random strategy allows for a good coverage, for instance the size of common vocabulary among the sub-corpora (after thresholding on frequency in each sub-corpora) is already more than $61\%$ of the top $300K$ vocabulary in the original corpus ( Note that this  corresponds to intersection of vocabularies of the sub-corpora and the union of vocabularies will be much larger).
In the next section, we elaborate the \mr framework used for training the sub-corpora in parallel and asynchronously. In the \mr framework we use a simple yet impressive (in terms of effectiveness) variant of \random strategy which we refer to as \shuffle. \shuffle also allows for a high coverage of vocabulary, for instance, the size of common vocabulary among sub-corpora is already $99.93\%$ of the vocabulary size in the original corpus.

\subsection{The Train Phase and the Shuffle Approach}
\label{sec:map-reduce}
    


        
    

We implement the sampling and training in a \mr framework to utilize parallel data loading and processing. Unlike earlier works~\cite{ordentlich2016network,stergiou2017distributed} that maintain and synchronize model state by expensive and frequent synchronization through message passing we employ a stateless approach for training.

In our approach the mappers are responsible for sampling the input into sub-corpora and the reducers are responsible for training. Say that we have a sampling rate $r\%$ thus needing to divide the corpus into $n = 100/r$ sub-corpora also resulting in $n$ corresponding models. We set the number of reducers to the number of models, i.e. $n$ reducers, with each reducer being responsible for training a model. The mappers implement the sampling by maintaining $n$ random number generators, one for each sub-corpora. For each input sentence we decide to assign it to a sub-corpora with a probability of $r/100$ (for each of the $n$ sub-corpora) and the sentence is then sent to corresponding reducer/s. Note that a sentence can be assigned to multiple sub-corpora. The $n$ reducers then train and generate a sub-model asynchronously on the sentences sent to them by the mappers. Training over multiple epochs involves multiple passes over the same training data. In distributed data processing frameworks like \mr this is typically realized using multiple rounds -- one for each epoch. 

\textbf{The Shuffle Approach.} To ensure that the same models receive exactly the same input, some amount of materialization of the samples is required (assuming the input is not memory resident) -- this is both wasteful and non-trivial. We instead propose a stateless approach called \shuffle where in each epoch (\mr round) we do not require that each sub-model receives exactly the same input sample as in the previous round/s. Note that each sub-model still receives the same fraction of the overall input data even if not the same training data. There are two distinct advantages to this. First, the training procedure is truly stateless and hence scalable. This is because we do not have to ensure the same training instances go to the same reducers which is typically implemented using content-based hashing techniques. 
\todo{Secondly, effectively a larger number of samples are trained as compared to \random which allows for high coverage of vocabulary resulting in better word representations. }
Secondly, and more importantly providing different samples across epochs to the same model has regularization effects and performs better than when the same input is seen across epochs (referred to as \random in our experiments). 

\subsection{The Merge Phase}
\label{sec:merge-phase}

In this section, we deal with the second part of our hypothesis, that is the merge phase, in which we are interested in finding a single embedding matrix (sometimes also known as the consensus embedding matrix) given $n$ $d$-dimensional word embedding matrices. Here we distinguish between two cases: (1) when all the $n$ sub-models have the same vocabulary and (2) when there exist words (present in atleast one sub-model) that are missing in one or more of the given matrices. 
\subsubsection{Case 1: Merging Sub-models with Common Vocabulary}
\label{sec:merging-intersection}
A very simple approach for case 1 is to concatenate the resulting matrices to obtain the final representation. Formally, let $V'$ be the collection of words that appear in all $n$ sub-models. For the $n$ trained sub-models with resulting word matrices 
$\mathbf{M}_1,...,\mathbf{M}_n$ (only for words in $V'$) each of dimension $|V'| \times d$, the simplest and surprisingly effective approach (as we will see in our experiments) is to simply concatenate the word matrices ${M}_{concat}$ of dimensions $|V'| \times d n$. We refer to this baseline as \textbf{Concat} or $\mathbf{M}_{concat} = 
\begin{bmatrix}
	\mathbf{M}_1|\mathbf{M}_2|\cdots|\mathbf{M}_n
\end{bmatrix}
$.
However, in the presence of a large number of sub-models the size of the merged matrix might already become large and hence undesirable. A simple alternative then is to employ Principle Component Analysis or PCA over the concatenated matrix and use a subset of the principle components as a representation, i.e.,  the first $d$ principal components of $M_{concat}$.
 We point to the reader that element wise averaging of the embedding vectors would not work unless the respective embeddings are first aligned. To understand this, consider the following case of 2 sub-models with 3 words represented by vectors  $\mathbf{w}_1^1=[1,1]$, $\mathbf{w}_1^2=[99,0]$, $\mathbf{w}_1^3=[1,-1]$ for the first sub-model and $\mathbf{w}_2^1=[-1,1]$, $\mathbf{w}_2^2=[-99,0]$, $\mathbf{w}_2^3=[-1, -1]$ corresponding to the second sub-model. In each sub-model word $1$ is most similar with word $3$, but in the averaged model ($\mathbf{w}^1=[0,1]$, $\mathbf{w}^2=[0,0]$, $\mathbf{w}^3=[0,-1]$) it is not the case. The problem is that the two embeddings are created independently and are in different spaces and need to be at first aligned. 
We next consider the case when one or more words are missing in some sub-models. We develop a variation of Generalized Procrustes Analysis (GPA) to deal with the missing vocabulary problem.
\subsubsection{Case 2: Merging Sub-models with Partial Vocabulary}
\label{sec:merging-missing}
The GPA approach to find consensus representation can be summarized as follows.

\begin{enumerate} 
\item Initialize the combined representation, say $Y$ by choosing randomly one of the representations or with random values.
\item Align all the representations to $Y$.
\item Calculate $Y'$ as the mean of the aligned representations.
\item If the error difference between $Y$ and $Y'$ is above a threshold, set $Y=Y'$ and return to step 2.
\end{enumerate}
Our proposed approach referred to as \problem (Alternating Linear Regression) follows the general principle of GPA with a novel adaptation to the case where some rows might be missing in some of the given representations.
Formally \problem learns a common representation $Y$ from multiple embeddings $\{ M_i\}$. Moreover, the vocabulary of $Y$ corresponds to the union of the vocabulary of all $M_i$. We describe the approach in the following.


\mpara{Initialization.}We experimented with two approaches for initialization: (i) randomly initialize all entries of $Y$ (ii) initialize the entries corresponding to the vocabulary intersection with the corresponding representations obtained by performing PCA over $M_{concat}$. 
During each iteration the algorithm then performs the following steps:
\begin{enumerate}
\item \textbf{Estimate Translation:} For each $M_i$ we first determine $M'_{i}$ (present) and $M^*_{i}$ (missing) corresponding to the vocabulary that is present or missing in model $M_{i}$. We denote the embeddings $Y'$ and $Y^*$ are the sub-matrices corresponding to the present and missing parts. 
We learn a transformation $W_i$ to aligm $M_i$ to $Y_i$ using the classical  Orthogonal Procrustes Analysis~\cite{schonemann1966generalized}.


\item \textbf{Estimating missing values:} We then estimate the corresponding $M^*_{i}$ by solving $Y^* = M^*_{i}W_{i}$ where we use $W_i$ and $Y^*$ from the previous steps.

\item \textbf{Update Joint Embedding:} Update $Y$ to be the mean of the translations of all $n$ models as follows and go to step 2.
Steps $1,2,3$ are repeated till the change in the average normalized Frobenius norm of displacement matrix, computed as
$\frac 1 n \sum_{i = 0}^r \frac {\norm{Y -  M_i W_i}_F} {\sqrt{|V| \cdot d}}$ will become smaller than a predefined threshold value.
\end{enumerate} 

%% file: evaluation.tex
\section{Experimental Setup}
\label{sec:exp-setup}
\begin{table}[ht!]
\centering
\footnotesize
\begin{tabular}{lllrr}
\hline
Benchmark & Task& Evaluation& \#unique & \#tests or\\
name& Type&Measure&words&  \#clusters\\
\hline
\men \cite{MENBruni2014multimodal}& Similarity& Spearman's $\rho$&751& 3000\\
\rg \cite{RG65Rubenstein1965contextual}& Similarity&Spearman's $\rho$&48& 65\\
\rw \cite{RWLuong-etal:conll13:morpho}& Similarity& Spearman's $\rho$&2951& 2034\\
\ws~\cite{WS353Finkelstein2001placing}& Similarity& Spearman's $\rho$&437& 353\\
\ap \cite{almuhareb2005concept}& Categorization& purity&402 & 21\\
\battig \cite{battig1969category}& Categorization& purity&4393 & 56\\
\google \cite{mikolov2013distributed}& Analogy& Accuracy&905& 19558\\
\semeval \cite{jurgens2012semeval}& Analogy& Accuracy&3224& 2531\\
\hline
\end{tabular}
\caption{Benchmarks statistics.}
\vspace{-8mm}
\label{tab:benchmarks}
\end{table}

\begin{table*}[ht!]
\centering
\footnotesize
\resizebox{\linewidth}{!}{
\begin{tabular}{lc|llllllll}
Division   & Sampling  & \ap & \battig & \men & \rg & \rw & \ws & \google & \semeval\\
Approach & Rate  & & & & & & & &\\
\hline
\partition  & 10\%    & \textbf{\underline{0.614}}  (0) &\textbf{\underline{0.450}} (210) &0.687 (0) &0.741 (0) & \textbf{\underline{0.374}}  (200) &0.636 (18) &0.533 (0) & 0.178 (61)\\
\random  & 10\%    &0.587 (0) &0.433 (217) &0.676 (0) &0.745 (0) &0.367 (270) &0.628 (18) &0.577 (0) & 0.182 (63)\\
\shuffle  & 10\%    &0.600 (2) & 0.447  (254) & \underline{0.712}  (0) & \textbf{\underline{0.781}}  (0) &0.331 (499) & \underline{0.651}  (18) & \textbf{\underline{0.657}} (1) & \textbf{\underline{0.185}} (82)\\
\hline
\partition  & 1\%    &0.488 (0) &0.370 (173) &0.393 (0) &0.508 (0) &0.288 (76) &0.378 (18) &0.239 (0) &0.162 (44)\\
\random  & 1\%    &0.512 (0) &0.363 (177) &0.410 (0) &0.530 (0) &0.280 (81) &0.372 (18) &0.267 (0) &0.162 (45)\\
\shuffle  & 1\%    &\underline{0.567} (2) & \underline{0.434} (254) & \underline{0.680} (0) & \underline{0.774} (0) & \underline{0.329} (499) & \underline{0.617} (18) & \underline{0.331} (1) & \underline{0.164} (82)\\
\hline
\baseline& -- &0.607 (0)& 0.442 (149)& \textbf{0.752} (0)& 0.731 (0)& 0.262 (54)& \textbf{0.666} (18)& 0.639 (0)& 0.175 (39)\\
\texttt{MLLib, 10 Cores}& -- &0.567 (2)& 0.407 (253)& 0.671 (0)& 0.691 (0)& 0.238 (487) & 0.567 (18)& 0.464 (1)& 0.152 (81)\\
\texttt{MLLib, 100 Cores}& -- &0.510 (2)& 0.366 (353)& 0.618 (0)& 0.670 (0)& 0.237 (487)& 0.578 (18)& 0.351 (1)& 0.130 (81)\\
\hline

\end{tabular}

}
\caption{Evaluation results for different sampling strategies. Merging using \problem initialized with \pca and run for 3 epochs. The numbers in parentheses indicate the count of words in each benchmark's vocabulary that is not present in the final merged model. Underlined values correspond to best result per sampling rate per benchmark. Bold values represent best result per benchmark.}
\vspace{-8mm}
\label{tab:sampling-comp}
\end{table*}

In order to establish the effectiveness and efficiency of our approach, we answer the following research questions in our experimental evaluation.

\begin{itemize}[leftmargin=*]
 \item \textbf{RQ I:} How does our proposed approach scale in terms of increasing data and increased parallelism? (Results in Section~\ref{sec:scaling}.) 
 \item \textbf{RQ II:} Which is the most effective sampling approach in partitioning the input corpus ?(Results in Section~\ref{sec:sampling}) 
 \item \textbf{RQ III:} What are the factors that determine the effectiveness of merging approaches and how do they handle sparsity and incompleteness in training ? (Results in Section~\ref{sec:merging}~\ref{sec:sparsity}.)
\end{itemize}
We first describe the experimental setup and materials used towards answering these research questions.

\subsection{Datasets}
\label{sec:datasets}



We used two large text datasets, \emph{Wikipedia} and \emph{Web}, in our experiments. Both corpora are pre-processed by removing non-textual elements, sentence splitting, and tokenization. 

\mpara{Wikipedia} refers to the English Wikipedia (August 2016 dump; uncompressed size = 14GB). The Wikipedia corpus contains $4,227,933$ sentences, spanning $2,313,580,449$ tokens. We use the Wikipedia corpus for all our effectiveness and scalability experiments. 

\mpara{Web} refers to a large text corpora of Web pages crawled in 2007 from the \texttt{.co.uk} domain. The dataset is 286 GB uncompressed. The Web corpus is far larger and contains $1,198,460,804$ sentences and $47,297,217,342$ tokens. We use the Web corpus only for our scalability experiments since building a baseline model for Web is computationally prohibitive. 

\mpara{Evaluation Benchmarks.} We test our models on a large set of benchmarks developed in the NLP community also suggested in \cite{jastrzebski2017evaluate}. Specifically, we select benchmarks to evaluate \emph{similarity, categorization and analogy} tasks (cf. Table~\ref{tab:benchmarks})  using word embeddings.

\subsection{Models Built} 
\label{sec:models}

We built the SGNS models using Gensim\footnote{https://radimrehurek.com/gensim/} framework ver. 3.4.0's word2vec implementation which trains representations using SGD and Hogwild~\cite{recht2011hogwild}. Gensim was configured to automatically use CPU-based BLAS\footnote{Using Intel\textsuperscript{\textregistered} Math Kernel Library} acceleration. 

For training, the window size is set to $10$, i.e. $10$ words to each side of the focus word. We fix the number of dimensions to $500$ for the \baseline baseline and the sub--models. For the \baseline and the \shuffle approach we set the size of vocabulary to $300,000$ (filtered by frequency) for both datasets before training. The vocabulary for \shuffle is precomputed and set in the first epoch. In the \partition and \random approaches, the word frequency threshold was set to $100/k$, where $k$ is the count of sub-models. This implies that for each sub-model only words that appear more than $100/k$ times are used in the vocabulary.
Furthermore, we created two more baseline models using Spark MLLib\footnote{Apache Spark 2.3.1 for Hadoop 2.6} with the same parameters as above, except for the vocabulary size which is only limited by a min. threshold of 100.

We also implemented Ordentlich et al. ~\cite{ordentlich2016network} for an additional baseline to
compare against. The implementation was done in Python\footnote{Running
all the computations via numpy using the Intel MKL BLAS implementation.},
but we didn't include any results since the runtime for just the 25\% subset of the Wikipedia
dataset was nearly 55 hours, making it unfeasible for larger data sizes.
Our implementation was spending most of the time running the actual computations
and waiting for the results of the shards.


The \baseline model is built on a single node with 256 GB of RAM and 2
Intel Xeon CPU E5-2620 v4 with 10 threads.  The $N$-divided sub-models and the \mllib models are trained on a compute cluster distributed using Cloudera 5.13.
Our compute cluster has 37 nodes: 10 nodes with 64GB RAM and 2x Intel Xeon CPU E5-2609 @ 2.40GHz,
13 nodes with 128GB RAM and 2x Intel Xeon CPU E5-2620 v2 @ 2.10GHz,
2 nodes with 128GB RAM and 2x Intel Xeon CPU E5-2620 v3 @ 2.40GHz and 
12 nodes with 256GB RAM and Intel Xeon CPU E5-2620 v4 @ 2.10GHz. All nodes are connected via Intel OmniPath with a theoretical max throughput of 58Gbps. We use 10 threads for the sub-models as well.

The models have an associated sampling rate  of $r\%$ which implies that the input is divided into $S=100/r$ samples. If $N$ is the total number of sentences, then each sample contains $N/S$ sentences.

\begin{table*}[ht]
\centering
\footnotesize
\resizebox{\linewidth}{!}{
\begin{tabular}{cc|llllllll}
 Sampling & Merging & \ap & \battig & \men & \rg & \rw & \ws & \google & \semeval\\
  Rate & Approach &  &  &  & & & & & \\
\hline
 10\% & \concat &0.614 (2)&0.435 (254)& \textbf{0.756} (0)&0.771 (0)&0.278 (499)&0.646 (18)& \textbf{0.674}  (1)& \textbf{0.190} (82)\\
 10\% & \pca & \textbf{0.654}  (2)& 0.452 (254)&0.719 (0)&0.786 (0)&0.329 (499)&0.650 (18)&0.652 (1)& 0.183 (82)\\
 10\% & \problem (\textsc{Rand}) &0.604 (2)& 0.441 (254)&0.715 (0)& 0.781 (0)& 0.329 (499)&\textbf{0.652} (18)&0.652 (1)& 0.183 (82)\\
 10\% & \problem (\textsc{PCA}) & 0.600 (2)&0.447 (254)&0.712 (0)&0.781 (0)&\textbf{0.331} (499)& 0.651 (18)&0.657 (1)& 0.185 (82)\\
 10\% & \textsc{Single Model} & 0.591 & 0.412 & 0.726 & 0.735 & 0.207 & 0.621 & 0.616 & 0.168\\
\hline
 5\% & \concat &0.602 (2)&0.435 (254)&0.748 (0)&0.763 (0)&0.272 (499)& 0.641 (18)&0.621 (1)& 0.184 (82)\\
 5\% & \pca &0.609 (2)&0.472 (254)&0.732 (0)&0.798 (0)&0.252 (499)&0.623 (18)&0.704 (1)& 0.182 (82)\\
 5\% & \problem  (\textsc{Rand}) & 0.609 (2)& \textbf{0.473}  (254)&0.729 (0)& \textbf{0.802}  (0)& 0.278 (499)&0.615 (18)&0.506 (1)& 0.182 (82)\\
 5\% & \problem (\textsc{PCA}) &0.631 (2)&0.479 (254)&0.729 (0)&0.803 (0)&0.280 (499)&0.617 (18)&0.517 (1)& 0.183 (82)\\
\hline
  1\% & \concat &0.560 (2)&0.399 (259)&0.711 (0)&0.776 (0)&0.244 (505)& 0.629 (18)& 0.472 (1)& 0.167 (84)\\
  1\% & \pca &0.550 (2)&0.419 (259)&0.683 (0)&0.777 (0)&0.252 (505)&0.616 (18)&0.348 (1)& 0.170 (84)\\
  1\% & \problem  (\textsc{Rand}) & 0.607 (2)& 0.440 (254)&0.680 (0)& 0.778 (0)& 0.329 (499)&0.616 (18)&0.329 (1)& 0.173 (82)\\
  1\% & \problem  (\textsc{Pca}) &0.567 (2)&0.434 (254)&0.680 (0)&0.774 (0)&0.329 (499)&0.617 (18)&0.331 (1)& 0.175 (82)\\
  1\% & \textsc{Single Model} & 0.481 & 0.346 & 0.528 & 0.685 & 0.187 & 0.422 & 0.040 & 0.120\\
  \hline
  &\baseline &0.607 (0)& 0.442 (149)& 0.752 (0)& 0.731 (0)& 0.262 (54)& 0.666 (18)& 0.639 (0)& 0.175 (39)\\
\hline

\end{tabular}
}
\caption{Evaluation results for different sampling rates and merging methods. The numbers in parentheses indicate the count of unique words in each benchmark's vocabulary that is not presented in the combined models or OOV terms. \problem (\textsc{Rand}) and \problem (\pca) correspond to \problem initialized by Random and \pca vectors respectively. }
\vspace{-5mm}
\label{tab:merging}
\end{table*}

\section{Experimental Results}
In this section, we finally provide answers to the research questions posed in the previous section. 
\label{sec:results}
\begin{figure}[ht]
	\centering
	\begin{subfigure}{\linewidth}
		\includegraphics[width=\textwidth]{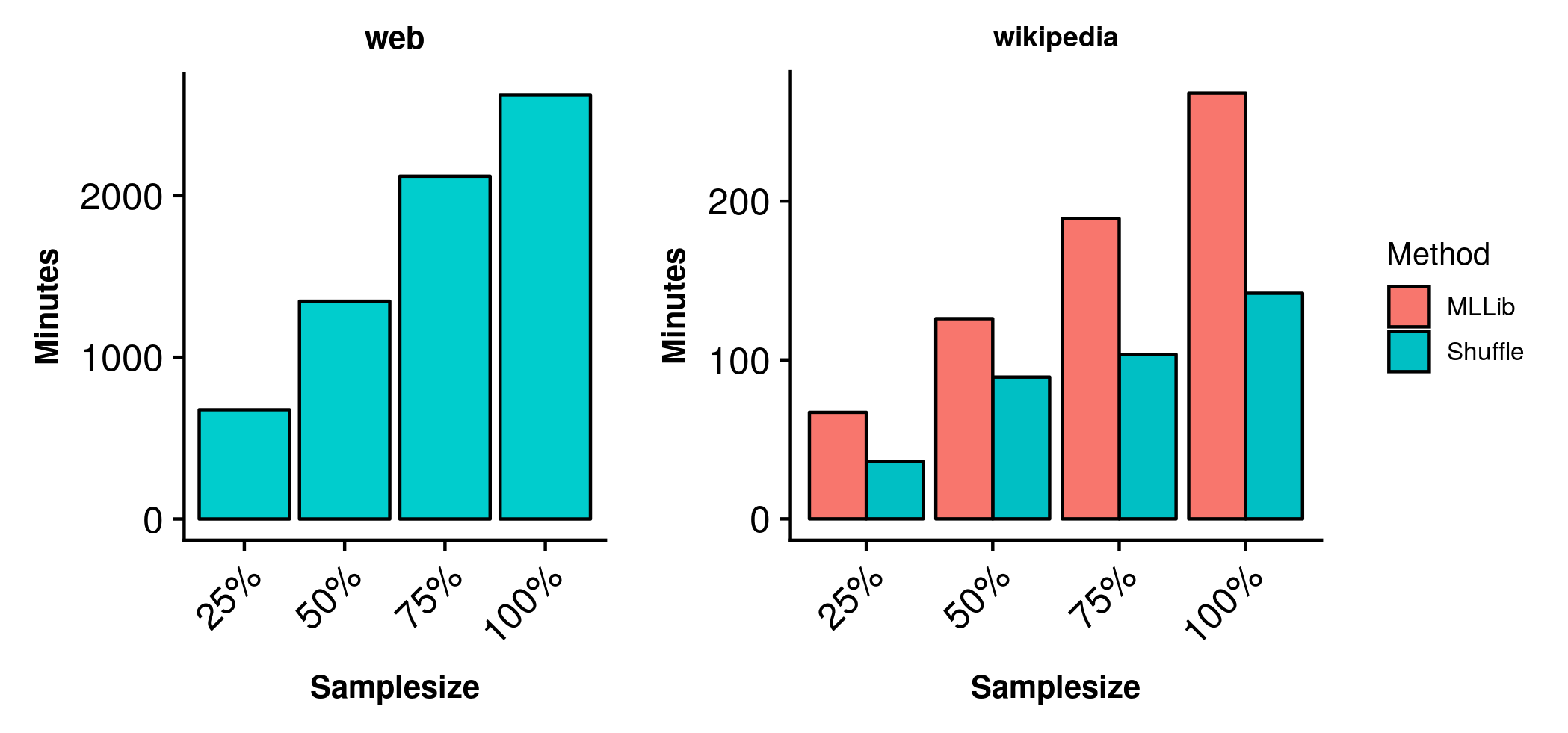}
	\end{subfigure}
    \caption{Time required for training increasing proportions of the Wikipedia and Web datasets with a 10\% sampling rate. Time for merging was omitted since they are negligible in comparison and too small to be visible.}
    \label{fig:data_time_scaling}
\end{figure}
\begin{table}[ht!]
\footnotesize
\centering
\begin{tabular}{l|rccc}
Sampling & Avg. Training   &  PCA & \problem \\
Rate& Time &   & (3 epochs)\\
\hline
1\% & 64 mins & 3.7 mins & 33.5 mins \\
5\% & 83 mins & 3.0 mins & 7.10 mins \\
6.67\% & 104 mins & 2.5 mins & 5.80 mins \\
10\% & 142 mins & 1.5 mins & 6.50 mins \\
20\% & 288 mins & 1.5 mins & 2.27 mins \\
25\% & 312 mins & 1.2 mins & 1.83 mins\\
33\% & 406 mins & 1.2 mins & 1.48 mins \\
50\% & 600 mins & 1.5 mins & 1.00 mins\\
\hline
\texttt{\baseline} & 1068 mins & -- & -- \\
\texttt{MLLib, 10 Cores} & 2146 mins & -- & -- \\
\texttt{MLLib, 100 Cores} & 268 mins & -- & -- \\
\hline
\end{tabular}
\caption{Wall-clock times for training and merging sub-models on the Wikipedia dataset with the \shuffle approach.}
\vspace{-10mm}
\label{tab:runtimes}
\end{table}

\subsection{Wallclock times}
\label{sec:scaling}

Firstly, we look at the training times of our approaches in
contrast to the time taken to train the \baseline model (Table~\ref{tab:runtimes}) as well as two \mllib models. 


We believe that optimizations relating to more efficient negative
sampling~\cite{ji2016parallelizing} are orthogonal to our
approaches and can be applied in a complementary manner. We observe that the \baseline model
takes ~17.8 hours to train for the Wikipedia dataset. The \mllib models only differ in the number of executors, 10 and 100, running for 35.8 and 4.5 hours respectively. In our executor set-up, we ensure that each model or sub-model is computed using an equal number of threads. For instance the 10 executor model corresponds to the \baseline model which uses 10 threads and the 100 executor model corresponds to the 10\% \shuffle model which is using 10 threads per sub--model.
Comparing training times, the 50\% \shuffle models
took an average training time of 600 minutes and the 10\%
\shuffle model took 142 minutes (as presented in
Table~\ref{tab:runtimes}). Firstly, this shows that SGNS
scales (almost) linearly to input size making it feasible to
train models in parallel in an asynchronous manner. We can also
see a decrease in the scaling performance when going to the 1\%
\shuffle model which took 64 minutes on average. This was most likely caused by using only 2 threads per reducer compared to the 10 threads per reducer for the other sampling rates which was necessary due to insufficient number of cores in the cluster. We also checked if we see the linear scaling for different data sizes for the 10\% \shuffle model.
This is still the case even for the larger web dataset as shown on the left hand side of Figure \ref{fig:data_time_scaling}. One can also observe a linear scaling for our approach on the smaller Wikipedia dataset, where we additionally compare the training time to the 100 executors \mllib model which needs approximately twice as much training time. Note that these times also take into account the map and shuffle
steps which have to be executed for each epoch. 

We now discuss the time taken to merge sub-models. We only show results for \pca and \problem since creating the concatenated model takes a negligible amount of time, and both depend on the concatenated model.

However \pca computation scales seamlessly with increasing number of sub-models (as shown in the table) with an increase of only 2.2 minutes when the number of sub-models increase from 10 to 100. \problem on the other hand takes a bit longer; merging 100 models takes 33.5 mins vs 6.5 mins when using \pca. Merging with \problem however has performance benefits as shown in section \ref{sec:merging}. When using a 10\% sampling rate, the merge times for both \pca and \problem are small when compared to the training time and performance is roughly the same in most benchmarks. For 1\% sampling however, the overall time taken is 67.7 minutes when merging with \pca and 97.5 minutes with \problem. While \problem takes nearly 50\% longer, it also performs better in 6/8 benchmarks (see Table~\ref{tab:merging}).

\subsection{Effect of Sampling}
\label{sec:sampling}

We begin by comparing the effect of two sampling approaches: \shuffle and \random. We contrast it with \partition for completeness in Table~\ref{tab:sampling-comp}. 
We fix the merging strategy as \problem (trained over 3 iterations, after which there is no change in performance) and report results for each of the sampling approaches.  We also experimented with two sampling rates -- $\{ 1\% , 10\% \}$ to check if different sample rates affects performance. 
We recall that the difference between \random and \shuffle approaches is that each reducer in the \shuffle approach might receive a different sample in each epoch whereas in \random the sample for each reducer is fixed and does not change with epochs.

The first observation we make is that \shuffle outperforms \random for 1\% sampling rate in all benchmarks and all except \rw for 10\% sampling rate. This establishes the superiority of \shuffle over \random and our intuition that using \shuffle (which uses different samples across epochs for the same sub-models) has a regularizing effect. Also, as expected \shuffle outperforms \partition in all the benchmarks with a sampling rate of 1\% consistently and by a large margin. For example, the gains in \men for 1\% sampling is almost 100\% and in \rg is 50\%. The gains are not as pronounced when a smaller sampling rate of 1\% is employed but still \shuffle is able to outperform \partition in all benchmarks. This also validates our justification of choosing random sampling and specifically \shuffle since it is able to preserve word and word-context distributions for each of the sub-models. Moreover, the common vocabulary in sub-models is much higher in case of \shuffle than \partition.

The \emph{main result} of this paper which also validates our Hypothesis is that merging larger sample sizes, i.e., \shuffle with 10\% sampling rate results in a performance that is either competitive or in most cases better than \baseline. With the exception of \men, we outperform the baseline in all benchmarks consistently. Sometimes \shuffle even with 1\% sampling rate outperforms the baseline (\rg). At the same time our approach is much faster than the \baseline as already discussed.

\subsection{Effect of Merging}
\label{sec:merging}

In the next set of experiments we present the results for merging the in-parallel computed sub-models. For this experiment we use the following merging techniques:

\begin{itemize}[leftmargin=*]
  \item Concatenating corresponding word vectors from each sample referred to as \concat. This approach is typically used as a baseline in many works~\cite{garten2015combining,goikoetxea2016single}.

  \item Principle Component Analysis (referred to as \pca) over the matrix formed from the concatenated vectors. 

  \item Our \problem approach that can either be initialized randomly denoted as \problem (\textsc{Rand}) or using the output of \pca i.e., \problem (\pca). 
  
 \item \textsc{Single Model} corresponds to using one sub-model instead of the merged model.
\end{itemize}

\begin{figure*}[t!]
\centering
     \begin{subfigure}{.45\linewidth}
     	\includegraphics[width=\linewidth]{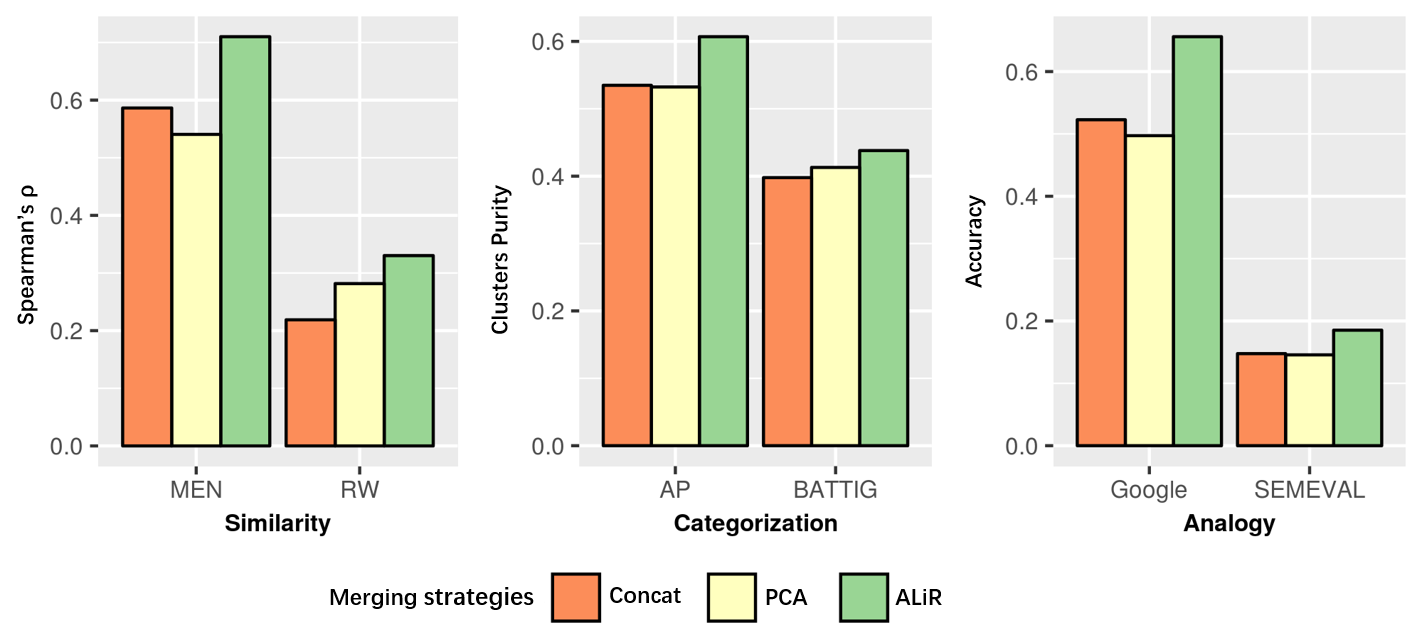}
        \caption{10\% words removed}
    \end{subfigure}
    \begin{subfigure}{.45\linewidth}
		\includegraphics[width=\linewidth]{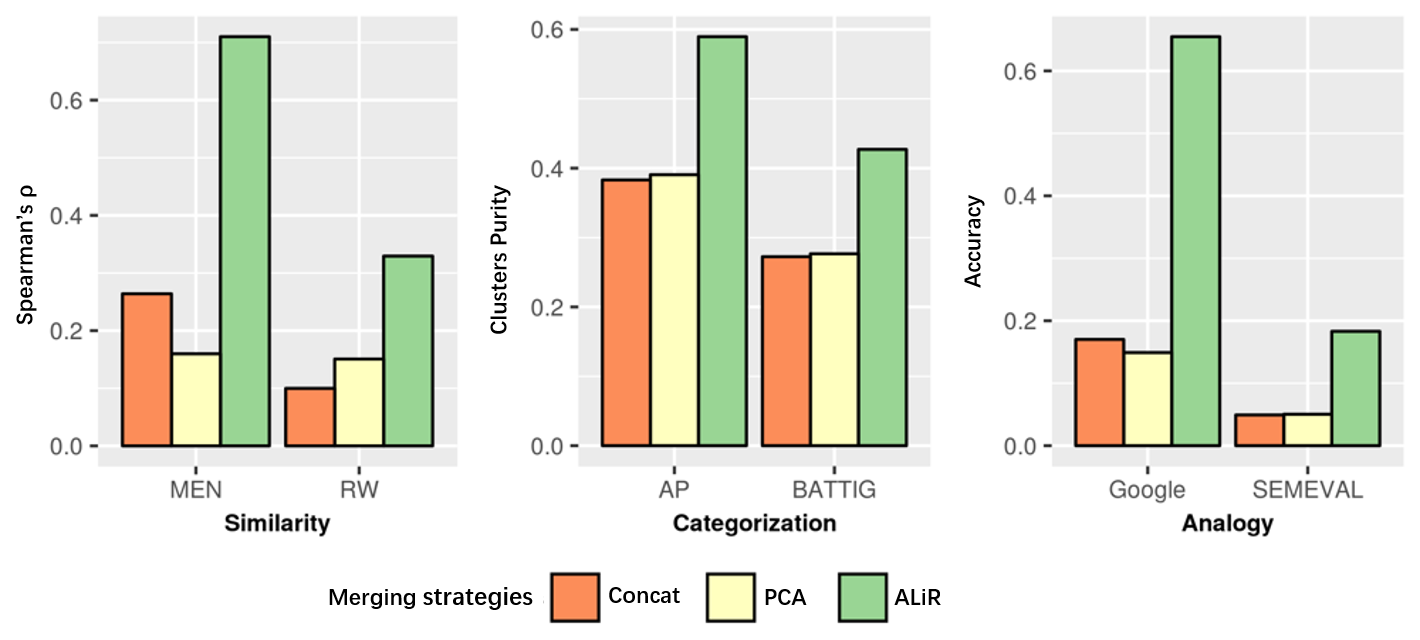}
        \caption{50\% words removed}
	\end{subfigure} 
    	\caption{Performance of missing-word reconstruction using \problem when compared to \concat and \pca. k\% of unique words from benchmark removed for at least one sub-model.}
        \label{fig:missing-words}
\end{figure*}


We also experimented with other complex dimensionality reduction approaches: low rank alignment or LRA~\cite{boucher2015aligning} and LLE~\cite{roweis2000nonlinear} but we did not see considerable improvements over simpler approaches like \pca. Additionally, these approaches are computationally more expensive and are prohibitive to employ when combining hundreds of models.  For \pca and \problem the target dimensionality is always set to be the same as the dimensionality of the \baseline  vectors. Since we established the superiority of \shuffle in the previous section we simply use \shuffle as the sampling approach. We consider 3 different sampling rates for these experiments.

The results are presented in Table~\ref{tab:merging}. We observe that \problem consistently performs best among its counterparts for the same sampling rates. Firstly, \problem outperforms \pca in 6/8 benchmarks for 1\% sampling rate and 4/8 for 10\% sampling rates. The closest competitor to \problem is unsurprisingly \concat that has a much higher dimensionality $d*n$ where $d$ is the dimensionality and $n$ is the number of sub-models. However, \concat is both impractical because of its high memory requirements and has a lower performance for a large number of smaller sub-models, i.e., 1\% sampling rate. The lower performance can be attributed to a decreasing number of terms that are indeed present in the vocabulary common to all sub-models. Secondly, as expected, the performance of models constructed using a higher sampling rate have a better performance as compared to those built using a smaller sampling rate.

One of the highlights in this experiment is that \problem with random initialization outperforms the baseline by $25\%$ on \rw. \rw is a difficult benchmark for embedding approaches in the sense that it has a high number of low frequency terms. Having a superior performance indicates the ability of \problem to be robust in OOV scenarios. 
We note here that merging approach has a clear advantage of just using a single sub-model in terms of vocabulary coverage. Moreover, our results  corresponding to using just one sub-model (averaged over single sub-models) as reported in Table~\ref{tab:merging} show significant gains achieved by the merged model.

In the next set of experiments we focus more on the scenario where there are more missing words in the sub-models and how well \problem reconstructs them. 




\subsection{Effect of Sparsity and Missing Vocabulary}
\label{sec:sparsity}


We assume that for the representation to be reconstructed, the word should be at least present in one of the sub-models. From the previous experimental setup, we found that the vocabulary covers nearly all unique words in most benchmarks. In Table~\ref{tab:merging}, we already see for \battig and \rw (where we have many OOV terms), \problem outperforms \concat and \pca across sampling rates. Also notice that the number of missing vocabulary terms is consistently lower or equal for \problem since our approach uses a union of the vocabulary terms whereas \concat and \pca take an intersection of vocabulary terms across sub-models. 

To study this effect more closely, we simulate the effect of more OOV terms by systematically removing benchmark words from sub-models. Figure~\ref{fig:missing-words} show the effect of removing 10\% and 50\% of the unique words from each benchmark for at least one sub-model. In these experiments, we randomly vary the number of sub-models from which we remove benchmark words, fix the sampling rate to 10\% and use \shuffle.  
From the results we notice that \problem outperforms \concat and \pca for all benchmarks since it can construct representations for missing words. \concat and \pca however ignore words not present in sub-models since no default vector is assumed for OOV words. This robustness to missing words is particularly strong when there are many OOV terms for the sub-models. For \ap, \men and \google, where we have nearly no OOV terms prior to removal, we observe that \problem does worse than \concat and \pca (Table~\ref{tab:merging}, sampling rate 10\%). However for the same benchmarks, when removing 10\% of benchmark words, \problem performs just as well whereas \concat and \pca perform significantly worse. 


This effect is more pronounced across all benchmarks when removing 50\% of the unique words. While \problem dips slightly in performance, \concat and \pca's performance drop is drastic (0.59 vs 0.27 for \concat, 0.57 vs 0.17 for \pca in \men). This result indicates that for collections with very large vocabularies, our parallel asynchronous training procedure with \problem can find good representations even if vocabularies are inconsistent across sub-models.

%% file: conclusions.tex
\section{Conclusions }
\label{sec:conclusion}
In this paper, we propose a scalable approach to train word embeddings by  partitioning the \emph{input space} in order to scale to massive text corpora while not sacrificing the performance of the embeddings. Our approach is simple, easy to implement and effective when compared against the baseline representation in a host of word similarity, analogy and categorization benchmarks. Our training procedure does not involve any parameter synchronization except a final sub-model merge phase that typically executes in a few minutes. 



%% file: main.bbl
\begin{thebibliography}{10}

\bibitem{almuhareb2005concept}
A.~Almuhareb and M.~Poesio.
\newblock Concept learning and categorization from the web.
\newblock In {\em Proceedings of the Cognitive Science Society}, volume~27,
  2005.

\bibitem{battig1969category}
W.~F. Battig and W.~E. Montague.
\newblock Category norms of verbal items in 56 categories a replication and
  extension of the connecticut category norms.
\newblock {\em Journal of experimental Psychology}, 80(3p2):1, 1969.

\bibitem{boehm2014hybrid}
M.~Boehm, S.~Tatikonda, B.~Reinwald, P.~Sen, Y.~Tian, D.~R. Burdick, and
  S.~Vaithyanathan.
\newblock Hybrid parallelization strategies for large-scale machine learning in
  systemml.
\newblock {\em Proceedings of the VLDB Endowment}, 7(7):553--564, 2014.

\bibitem{bojanowski2016enriching}
P.~Bojanowski, E.~Grave, A.~Joulin, and T.~Mikolov.
\newblock Enriching word vectors with subword information.
\newblock {\em arXiv preprint arXiv:1607.04606}, 2016.

\bibitem{boucher2015aligning}
T.~Boucher, C.~Carey, S.~Mahadevan, and M.~D. Dyar.
\newblock Aligning mixed manifolds.
\newblock In {\em AAAI}, pages 2511--2517, 2015.

\bibitem{MENBruni2014multimodal}
E.~Bruni, N.-K. Tran, and M.~Baroni.
\newblock Multimodal distributional semantics.
\newblock {\em J. Artif. Intell. Res.(JAIR)}, 49(2014):1--47, 2014.

\bibitem{cho2014learning}
K.~Cho, B.~{Van Merri{\"{e}}nboer}, C.~Gulcehre, D.~Bahdanau, F.~Bougares,
  H.~Schwenk, and Y.~Bengio.
\newblock {Learning phrase representations using RNN encoder-decoder for
  statistical machine translation}.
\newblock {\em arXiv preprint arXiv:1406.1078}, 2014.

\bibitem{WS353Finkelstein2001placing}
L.~Finkelstein, E.~Gabrilovich, Y.~Matias, E.~Rivlin, Z.~Solan, G.~Wolfman, and
  E.~Ruppin.
\newblock Placing search in context: The concept revisited.
\newblock In {\em Proceedings of the 10th international conference on World
  Wide Web}, pages 406--414. ACM, 2001.

\bibitem{garten2015combining}
J.~Garten, K.~Sagae, V.~Ustun, and M.~Dehghani.
\newblock Combining distributed vector representations for words.
\newblock In {\em VS@ HLT-NAACL}, pages 95--101, 2015.

\bibitem{ghannay2016word}
S.~Ghannay, B.~Favre, Y.~Esteve, and N.~Camelin.
\newblock Word embedding evaluation and combination.
\newblock In {\em LREC}, 2016.

\bibitem{ghoting2011systemml}
A.~Ghoting, R.~Krishnamurthy, E.~Pednault, B.~Reinwald, V.~Sindhwani,
  S.~Tatikonda, Y.~Tian, and S.~Vaithyanathan.
\newblock Systemml: Declarative machine learning on mapreduce.
\newblock In {\em Data Engineering (ICDE)}, pages 231--242. IEEE, 2011.

\bibitem{goikoetxea2016single}
J.~Goikoetxea, E.~Agirre, and A.~Soroa.
\newblock Single or multiple? combining word representations independently
  learned from text and wordnet.
\newblock In {\em AAAI}, pages 2608--2614, 2016.

\bibitem{Gower1975}
J.~C. Gower.
\newblock Generalized procrustes analysis.
\newblock {\em Psychometrika}, 40(1):33--51, 1975.

\bibitem{jastrzebski2017evaluate}
S.~Jastrzebski, D.~Le{\'s}niak, and W.~M. Czarnecki.
\newblock How to evaluate word embeddings? on importance of data efficiency and
  simple supervised tasks.
\newblock {\em arXiv preprint arXiv:1702.02170}, 2017.

\bibitem{ji2016parallelizing}
S.~Ji, N.~Satish, S.~Li, and P.~Dubey.
\newblock Parallelizing word2vec in shared and distributed memory.
\newblock {\em arXiv preprint arXiv:1604.04661}, 2016.

\bibitem{jurgens2012semeval}
D.~A. Jurgens, P.~D. Turney, S.~M. Mohammad, and K.~J. Holyoak.
\newblock Semeval-2012 task 2: Measuring degrees of relational similarity.
\newblock In {\em Proceedings of the First Joint Conference on Lexical and
  Computational Semantics}, pages 356--364, 2012.

\bibitem{levy2014neural}
O.~Levy and Y.~Goldberg.
\newblock {Neural Word Embedding as Implicit Matrix Factorization}.
\newblock {\em Advances in Neural Information Processing Systems (NIPS)}, pages
  2177--2185, 2014.

\bibitem{li2015word}
Y.~Li, L.~Xu, F.~Tian, L.~Jiang, X.~Zhong, and E.~Chen.
\newblock Word embedding revisited: A new representation learning and explicit
  matrix factorization perspective.
\newblock In {\em IJCAI}, pages 3650--3656, 2015.

\bibitem{RWLuong-etal:conll13:morpho}
M.-T. Luong, R.~Socher, and C.~D. Manning.
\newblock Better word representations with recursive neural networks for
  morphology.
\newblock In {\em CoNLL}, Sofia, Bulgaria, 2013.

\bibitem{mikolov2013distributed}
T.~Mikolov, I.~Sutskever, K.~Chen, G.~S. Corrado, and J.~Dean.
\newblock Distributed representations of words and phrases and their
  compositionality.
\newblock In {\em Advances in neural information processing systems}, pages
  3111--3119, 2013.

\bibitem{mikolov2013linguistic}
T.~Mikolov, W.-t. Yih, and G.~Zweig.
\newblock Linguistic regularities in continuous space word representations.
\newblock In {\em hlt-Naacl}, volume~13, pages 746--751, 2013.

\bibitem{mitra2017neural}
B.~Mitra and N.~Craswell.
\newblock Neural models for information retrieval.
\newblock {\em arXiv preprint arXiv:1705.01509}, 2017.

\bibitem{muromagi2017linear}
A.~Murom{\"a}gi, K.~Sirts, and S.~Laur.
\newblock Linear ensembles of word embedding models.
\newblock {\em arXiv preprint arXiv:1704.01419}, 2017.

\bibitem{desm16}
E.~Nalisnick, B.~Mitra, N.~Craswell, and R.~Caruana.
\newblock Improving document ranking with dual word embeddings.
\newblock In {\em Proceedings of the 25th International Conference Companion on
  World Wide Web}, pages 83--84, 2016.

\bibitem{ordentlich2016network}
E.~Ordentlich, L.~Yang, A.~Feng, P.~Cnudde, M.~Grbovic, N.~Djuric,
  V.~Radosavljevic, and G.~Owens.
\newblock Network-efficient distributed word2vec training system for large
  vocabularies.
\newblock In {\em Proceedings of the 25th ACM International on Conference on
  Information and Knowledge Management}, pages 1139--1148. ACM, 2016.

\bibitem{Pennington2014glove}
J.~Pennington, R.~Socher, and C.~D. Manning.
\newblock {GloVe: Global Vectors for Word Representation}.
\newblock {\em Proceedings of the 2014 Conference on Empirical Methods in
  Natural Language Processing}, pages 1532--1543, 2014.

\bibitem{recht2011hogwild}
B.~Recht, C.~Re, S.~Wright, and F.~Niu.
\newblock Hogwild: A lock-free approach to parallelizing stochastic gradient
  descent.
\newblock In {\em Advances in Neural Information Processing Systems}, pages
  693--701, 2011.

\bibitem{roweis2000nonlinear}
S.~T. Roweis and L.~K. Saul.
\newblock Nonlinear dimensionality reduction by locally linear embedding.
\newblock {\em science}, 290(5500):2323--2326, 2000.

\bibitem{RG65Rubenstein1965contextual}
H.~Rubenstein and J.~B. Goodenough.
\newblock Contextual correlates of synonymy.
\newblock {\em Communications of the ACM}, 8(10):627--633, 1965.

\bibitem{schonemann1966generalized}
P.~H. Sch{\"o}nemann.
\newblock A generalized solution of the orthogonal procrustes problem.
\newblock {\em Psychometrika}, 31(1):1--10, 1966.

\bibitem{socher2013parsing}
R.~Socher, J.~Bauer, C.~D. Manning, et~al.
\newblock Parsing with compositional vector grammars.
\newblock In {\em Proceedings of the 51st Annual Meeting of the Association for
  Computational Linguistics (Volume 1: Long Papers)}, volume~1, pages 455--465,
  2013.

\bibitem{socher2013recursive}
R.~Socher, A.~Perelygin, J.~Wu, J.~Chuang, C.~D. Manning, A.~Ng, and C.~Potts.
\newblock Recursive deep models for semantic compositionality over a sentiment
  treebank.
\newblock In {\em Proceedings of the 2013 conference on empirical methods in
  natural language processing}, pages 1631--1642, 2013.

\bibitem{DBLP:journals/corr/SpeerC16}
R.~Speer and J.~Chin.
\newblock An ensemble method to produce high-quality word embeddings.
\newblock {\em CoRR}, abs/1604.01692, 2016.

\bibitem{stergiou2017distributed}
S.~Stergiou, Z.~Straznickas, R.~Wu, and K.~Tsioutsiouliklis.
\newblock Distributed negative sampling for word embeddings.
\newblock In {\em AAAI}, pages 2569--2575, 2017.

\bibitem{ustun2014distributed}
V.~Ustun, P.~S. Rosenbloom, K.~Sagae, and A.~Demski.
\newblock Distributed vector representations of words in the sigma cognitive
  architecture.
\newblock In {\em International Conference on Artificial General Intelligence},
  pages 196--207. Springer, 2014.

\bibitem{eickhoff2016efficient}
J.~B.~P. Vuurens, C.~Eickhoff, and A.~P. de~Vries.
\newblock Efficient parallel learning of word2vec.
\newblock {\em ICML 2016 Machine Learning workshop}, 48, 2016.

\bibitem{weston2011wsabie}
J.~Weston, S.~Bengio, and N.~Usunier.
\newblock Wsabie: Scaling up to large vocabulary image annotation.
\newblock 2011.

\bibitem{xing2015petuum}
E.~P. Xing, Q.~Ho, W.~Dai, J.~K. Kim, J.~Wei, S.~Lee, X.~Zheng, P.~Xie,
  A.~Kumar, and Y.~Yu.
\newblock Petuum: A new platform for distributed machine learning on big data.
\newblock {\em IEEE Transactions on Big Data}, 1(2):49--67, 2015.

\bibitem{yin2015learning}
W.~Yin and H.~Sch{\"u}tze.
\newblock Learning meta-embeddings by using ensembles of embedding sets.
\newblock {\em arXiv preprint arXiv:1508.04257}, 2015.

\bibitem{zaharia2010spark}
M.~Zaharia, M.~Chowdhury, M.~J. Franklin, S.~Shenker, and I.~Stoica.
\newblock Spark: Cluster computing with working sets.
\newblock {\em HotCloud}, 10(10-10):95, 2010.

\bibitem{zamani2016estimating}
H.~Zamani and W.~B. Croft.
\newblock Estimating embedding vectors for queries.
\newblock In {\em Proceedings of the 2016 ACM on International Conference on
  the Theory of Information Retrieval}, pages 123--132. ACM, 2016.

\bibitem{inproc:Zhao2015}
K.~Zhao, H.~Hassan, and M.~Auli.
\newblock Learning translation models from monolingual continuous
  representations.
\newblock In {\em Proceedings of the 2015 Conference of the North American
  Chapter of the Association for Computational Linguistics: Human Language
  Technologies}, pages 1527--1536, 2015.

\end{thebibliography}
